\documentclass[runningheads]{llncs}
\usepackage[T1]{fontenc}
\usepackage{graphicx}
\usepackage{hyperref}
\hypersetup{
    colorlinks=true,
    linkcolor=red,
    urlcolor=magenta,
    citecolor=blue
}

\usepackage{amsmath}
\usepackage{amssymb}
\usepackage{algpseudocode}
\usepackage{algorithm}
\usepackage{booktabs}
\usepackage{dblfloatfix}
\usepackage{cite}
\usepackage[dvipsnames]{xcolor}
\usepackage{soul} 
\newcommand\blfootnote[1]{
    \begingroup
    \renewcommand\thefootnote{}\footnote{#1}
    \addtocounter{footnote}{-1}
    \endgroup
}

\begin{document}

\title{GP-PCS: One-shot Feature-Preserving Point Cloud Simplification with Gaussian Processes on Riemannian Manifolds
}
%
%
\author{Stuti Pathak\inst{1}$^\dagger$\orcidID{0009-0001-8973-4822} \and
Thomas Baldwin-McDonald\inst{2}$^\dagger$\orcidID{0000-0001-7301-4399} \and
Seppe Sels\inst{1}\orcidID{0000-0002-0590-2770} \and
Rudi Penne\inst{1}\orcidID{0000-0002-0921-1950}}
\authorrunning{S. Pathak et al.}
%
\institute{University of Antwerp, Prinsstraat 13, Antwerp, 2000, Belgium\and
University of Manchester, Oxford Road, Manchester, M13 9PL, United Kingdom}
\maketitle              
\begin{abstract}
The processing, storage and transmission of large-scale point clouds is an ongoing challenge in the computer vision community which hinders progress in the application of 3D models to real-world settings, such as autonomous driving, virtual reality and remote sensing. We propose a novel, one-shot point cloud simplification method which preserves both the salient structural features and the overall shape of a point cloud without any prior surface reconstruction step. Our method employs Gaussian processes suitable for functions defined on Riemannian manifolds, allowing us to model the surface variation function across any given point cloud. A simplified version of the original cloud is obtained by sequentially selecting points using a greedy sparsification scheme. The selection criterion used for this scheme ensures that the simplified cloud best represents the surface variation of the original point cloud. We evaluate our method on several benchmark and self-acquired point clouds, compare it to a range of existing methods, demonstrate its application in downstream tasks of registration and surface reconstruction, and show that our method is competitive both in terms of empirical performance and computational efficiency. The code is available at \href{https://github.com/stutipathak5/gps-for-point-clouds}{https://github.com/stutipathak5/gps-for-point-clouds}.

\end{abstract}
\section{Introduction}
\label{intro}
\blfootnote{$^\dagger$ Equal contribution.}

Recent years have seen a growing need for the conversion of real-world objects to computerized models \cite{xiao2021estimating, fernandes2021point} across several domains, such as digital preservation of cultural heritage \cite{PIERACCINI200163} and manufacturing of mechanical parts for industry \cite{LI200253}. This need has given rise to a range of modern data acquisition techniques such as laser scanning, which densely samples the surface of a 3D object, thereby generating millions of significantly redundant data points. 3D models can be obtained from this \textit{point cloud} by constructing a polygonal mesh using techniques such as the \textit{ball-pivoting algorithm} and \textit{Poisson surface reconstruction} \cite{bernardini1999ball, kazhdan2006poisson, berger2014state}. However, the sheer size of these dense point clouds makes this task computationally expensive in terms of both memory and time. Furthermore, the size of such generated meshes impedes further processing efforts, and necessitates the use of costly mesh simplification strategies \cite{garland1997surface, hoppe1993mesh, cignoni1998comparison} for size reduction. This makes efficient simplification of the underlying point cloud, prior to any surface reconstruction, an important and impactful problem which if addressed, has the potential to significantly improve the scalability of several computer vision applications. 

The inherent dependency of surface reconstruction methods on surface normals, makes the visual perceptual quality of a point cloud an indirect yet important aspect of any mesh processing pipeline \cite{ cignoni1998comparison}. Although it is difficult to quantify this visual degradation in the case of point cloud simplification methods, one can say that the more enhanced the characteristic features of an object (such as sharp edges and high curvature regions) are in the simplified cloud, the higher is its human perceptual quality \cite{lee2005mesh}. Therefore, an optimal point cloud simplification technique should preserve both the global structural appearance, and the salient features of the point cloud in question. Some of these methods will be discussed in detail in the upcoming section.

Given that the point cloud representing an object exists on a Riemannian manifold in 3D space, Euclidean distance fails to measure the intrinsic distance between any two points on its surface. Recently, techniques which extend existing machine learning methods to model functions defined on manifolds have gained popularity.
For instance, \textit{Gaussian
processes} (GPs), a widely used class of non-parametric statistical
models, which often use Euclidean distance-based
covariance functions, have been made compatible for functions
whose domains are compact Riemannian manifolds
using ideas from harmonic analysis \cite{borovitskiy2020matern}.

\vspace{-0.5cm}
\begin{figure}
     \centering 
        \includegraphics[scale=0.58, trim=4.15cm 11cm 5cm 1.5cm, clip]{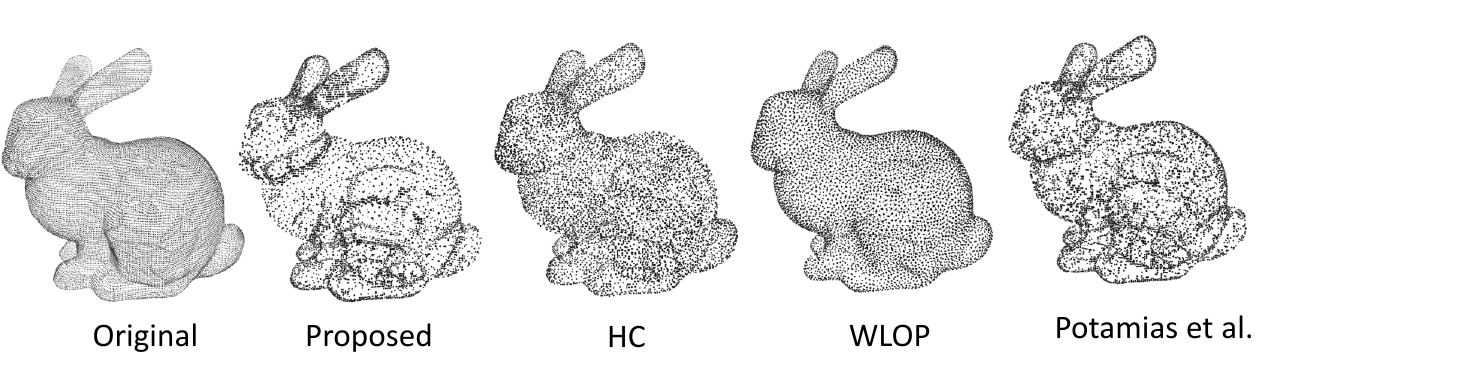} \vspace{-0.5cm}
         \caption{Point cloud simplification methods typically fail to strike a balance between preserving sharp features and maintaining the overall structure of the original cloud. Our approach circumvents this trade-off by achieving both targets, as is evident from the simplified versions of the Stanford Bunny \cite{levoy2005stanford} 
         obtained using the proposed technique and three pre-existing methods; Hierarchical Clustering (HC) \cite{pauly2002efficient}, Weighted Locally
Optimal Projection (WLOP) \cite{huang2009consolidation}, and Potamias \textit{et al.} \cite{potamias2022revisiting} simplification.}
         \label{fig:figure1}
\end{figure}
\vspace{-0.5cm}
In this work, we propose a novel, one-shot, feature-preserving simplification method using GPs with kernels defined on Riemannian manifolds. Using a greedy algorithm for GP sparsification, we iteratively construct a simplified representation of a point cloud without the need for any prior surface reconstruction or training on large point cloud datasets. We experiment on several point clouds, compare with several techniques and demonstrate competitive results both empirically and in terms of computational efficiency. Qualitatively, as shown in Figure \ref{fig:figure1}, our method effectively preserves visual features whilst providing a sufficiently dense coverage of the domain of the original cloud.

\subsubsection{Outline of the paper:} Section \ref{sec:rw} briefly reviews a number of existing point cloud simplification techniques which are relevant to our work. Section \ref{sec:background} provides background details regarding the computation of surface variation, GPs with kernels defined on non-Euclidean domains and a greedy subset-of-data scheme for GP inference. Section \ref{sec:methodology} outlines the proposed GP-based point cloud simplification algorithm. Section \ref{sec:experiments}, in combination with the supplementary material, includes an empirical evaluation of our method on various benchmark and self-acquired point clouds, with comparisons to competing simplification techniques, along with applications to some downstream tasks and ablation studies. Finally, Section \ref{sec:conc} summarises our contributions and provides a brief discussion of the scope for future work.

\section{Related work}
\label{sec:rw}
In this section we will introduce a number of existing point cloud simplification techniques, with a particular focus on works which have a feature-preserving element to their approach. Some of the earliest curvature-sensitive simplification techniques were proposed by Pauly \textit{et al.} \cite{pauly2002efficient} and Moenning \textit{et al.} \cite{moenning2003new}. The former method, termed \textit{Hierarchical Clustering} (HC), recursively divides the original point cloud into two sets, until each child set attains a size smaller than a threshold \textit{size parameter}. Moreover, a \textit{variation parameter} plays an important role in sparsifying regions of low curvature by selective splitting. The perceptual quality and the size of the simplified cloud depend entirely on these two parameters, which must be carefully and manually tuned, making HC unsuitable for automated applications. Additionally, the surface reconstructions obtained from HC-simplified point clouds are often poor for clouds with complex surfaces, as will be seen in Section \ref{sec:experiments}. This is because it is challenging to tune the parameters of HC in such a way that preservation of sharp features is achieved whilst still ensuring dense coverage of the original cloud. 

Another widely-used technique is \textit{Weighted Locally Optimal Projection} (WLOP) proposed by Huang \textit{et al.} \cite{huang2009consolidation}. In this work, the authors modified the existing parameterization-free denoising simplification scheme termed \textit{Locally Optimal Projection} (LOP) \cite{lipman2007parameterization}, which is  unsuitable for non-uniformly distributed point clouds. WLOP overcomes this limitation by incorporating locally adaptive density weights into LOP. Although WLOP results in an evenly distributed simplified cloud, it still lacks sensitivity towards salient geometric features which will also become apparent in Section \ref{sec:experiments}. Recently, Potamias \textit{et al.} \cite{potamias2022revisiting} have proposed a graph neural network-based learnable simplification technique which uses a modified variant of Chamfer distance in order to backpropagate errors. Their method can simplify point clouds in real-time but involves a computationally intensive training process using large point cloud datasets such as TOSCA \cite{bronstein2008numerical}. Moreover, their model's efficiency is limited to simplifying point clouds which are structurally similar to the learned data, as inherently neural networks struggle to generalize outside of the domain of the training data. Even more recent work from Wu \textit{et al.} \cite{wu2023attention}, named \textit{APES}, proposes an edge-sampling method which claims to capture the salient points within a point cloud using an attention mechanism. As shown in their paper, this technique generally provides good results for some point cloud tasks. However, as discussed by the authors themselves, the edge-enhancing nature of their method hinders upsampling operations, which can lead to poor reconstruction and segmentation results later.

\textit{Approximate Intrinsic Voxel Structure for Point Cloud Simplification} (AIVS), introduced by Lv \textit{et al.} \cite{lv2021approximate}, combines global voxel structure and local farthest point sampling to generate simplification demand-specific clouds which can be either isotropic, curvature-sensitive or have sharp edge preservation. As with HC however, AIVS requires manual tuning of user-specified parameters in order to obtain optimal results. Additionally, even in parallel computation mode, AIVS is quite costly in terms of computational runtime. Potamias \textit{et al.} and  Lv \textit{et al.} do not provide open-source implementations of their curvature-sensitive simplification techniques, which poses a challenge for reproducibility and benchmarking. However, we thank the authors of Potamias \textit{et al.} for directly providing some simplified point clouds; their results are included later in this paper. Qi \textit{et al.} \cite{qi2019feature} introduced \textit{PC-Simp}, a method which aims to produce uniformly-dense and feature-sensitive simplified clouds, leveraging ideas from graph signal processing. This uniformity depends on a \textit{weight parameter} which as with HC and AIVS, is user-specified. Alongside simplification, they also apply their technique to point cloud registration. However, in practice PC-Simp is unreliable for complex-surfaced point clouds as it fails to provide a high degree of feature-preservation, regardless of the weight parameter chosen. Additionally, as discussed later in Section \ref{sec:experiments}, the runtime of this technique is considerably longer than any other method tested.

Finally, it has been observed that most of the aforementioned works on feature-preserving point cloud simplification schemes experiment on structurally simple point clouds. Furthermore, surface reconstruction results are rarely presented and discussed. Hence, to underline the efficiency of our method, we experiment on point clouds generated from complex-surfaced objects and provide the corresponding reconstruction results. Also, some of the datasets used by the mentioned techniques are synthetically generated and already have a higher concentration of points around salient features when compared to low curvature regions (for example, the TOSCA dataset). Hence, unlike them, we do not experiment on point clouds from these datasets as it defeats the purpose of being a feature-sensitive simplification technique.

\section{Background}
\label{sec:background}

\subsection{Surface variation}
 \label{sec:surface}
 
Consider an unstructured dense point cloud $P=\{\boldsymbol{p_{1}}, \boldsymbol{p_{2}}, ..., \boldsymbol{p_{N}}\}$ of size $N$ existing in 3D Euclidean space, $\mathbb{R}^3$. We can generate the local neighbourhood $N_{\boldsymbol{p_i}}$ of each point $\boldsymbol{p_{i}}$ in $P$ by two different methods. Firstly, we can gather all of the points within a certain Euclidean distance $r$ from $\boldsymbol{p_i}$; this approach is referred to as \textit{radius search}. Alternatively, we can gather all of the k-nearest Euclidean neighbours of $\boldsymbol{p_i}$, which is referred to as \textit{KNN search}. The choice of this scale-factor ($r$ or $k$) not only depends on the size and density of a point cloud but also on the desired level of detail for a given application. These aspects make the task of automatic estimation of the neighbourhood of a point in a cloud an important, yet challenging one \cite{rusu2010semantic}. In this work, we implement the approach taken by the \textit{CloudCompare} software package, 
where this process is automated by first calculating an approximate surface per point from the bounding box volume. This estimated value, along with a user-defined approximate neighbour number, is used to estimate a radius $r$, which is then used to perform radius search for each point. In our method, we have fixed this approximate neighbour number to $25$ as it provides good empirical performance across a wide variety of point clouds. However, we provide ablation studies over a range of neighbourhood sizes in the supplementary material.

Several local surface properties \cite{thomas2018semantic} of the point cloud at a given query point $\boldsymbol{p_i}$ can be estimated by analysing the eigenvalues and eigenvectors of the covariance matrix $\mathbf{C_i}$ defined by the point's neighbourhood $N_{\boldsymbol{p_i}}=\{\boldsymbol{p_{i_1}}, \boldsymbol{p_{i_2}}, ..., \boldsymbol{p_{i_n}}\}$: 
\begin{equation} \label{}
\mathbf{C_i}={\begin{bmatrix}
\boldsymbol{p_{i_1}}-\boldsymbol{\bar{p_i}} \\
\boldsymbol{p_{i_2}}-\boldsymbol{\bar{p_i}} \\
... \\
\boldsymbol{p_{i_n}}-\boldsymbol{\bar{p_i}}
\end{bmatrix}}^T 
\cdot 
\begin{bmatrix}
\boldsymbol{p_{i_1}}-\boldsymbol{\bar{p_i}} \\
\boldsymbol{p_{i_2}}-\boldsymbol{\bar{p_i}} \\
... \\
\boldsymbol{p_{i_n}}-\boldsymbol{\bar{p_i}}
\end{bmatrix},
\end{equation}
where, $\boldsymbol{\bar{p_i}}$ is the centroid of all the points $\boldsymbol{p_{i_i}} \in N_{\boldsymbol{p_i}}$. By means of \textit{principal component analysis} (PCA), we may now fit a plane tangent to the 3D surface, formed by all of the points within $N_{\boldsymbol{p_i}}$, at $\boldsymbol{\bar{p_i}}$. As $\mathbf{C_i}$ is a $3\times3$ symmetric and positive semi-definite matrix, all of its eigenvalues $\left(\lambda_j, j\in\{0,1,2\}\right)$ are positive and real, whilst the corresponding eigenvectors ($\boldsymbol{v_j}$) form an orthogonal frame corresponding to the principal components of $N_{\boldsymbol{p_i}}$. If $0\leq\lambda_0\leq\lambda_1\leq\lambda_2$, then $\boldsymbol{v_2}$ and $\boldsymbol{v_1}$ span the aforementioned tangent plane, whilst $\boldsymbol{v_0}$ represents the vector perpendicular to it. Therefore, $\boldsymbol{v_0}$ can be considered as an estimate of the surface normal to the point cloud (without actual surface reconstruction) at query point $\boldsymbol{p_i}$. Furthermore, as defined by Pauly \textit{et al.} \cite{pauly2002efficient}, we can calculate the \textit{surface variation} at the query point as:
\begin{equation} \label{sf}
\sigma_n(\boldsymbol{p_i})=\frac{\lambda_0}{\lambda_0+\lambda_1+\lambda_2}.
\end{equation}
This quantity is not only closely related to the surface curvature at $\boldsymbol{p_i}$ but also serves as a more suitable criterion for simplification, as discussed in detail by the authors \cite{pauly2002efficient}.
%
\subsection{Gaussian processes on Riemannian manifolds}
\label{sec:riemannian_gp}
Gaussian processes (GPs) are non-parametric Bayesian models which allow for a rigorous estimation of predictive uncertainty, and have been widely studied and applied by the machine learning community over the last two decades. Consider a scenario where we have a training dataset of $N$ observations, $\{\mathbf{x}_i, y_i\}_{i=1}^N$, where $\mathbf{x}_i \in \mathbb{R}^P$ and $y_i \in \mathbb{R}$. In our application, $\mathbf{x}_i \in \mathbb{R}^3$ is a Euclidean coordinate, and $y_i$ is the surface variation associated with said coordinate. We assume access to noisy observations of an underlying latent function, such that $y_i = f(\mathbf{x}_i) + \epsilon_i$, where $\epsilon_i \sim \mathcal{N}(0, \sigma_y^2)$. A GP defines a distribution over functions which we can use to infer the form of the true latent function which generated our training data. The GP prior can be written as
$f \sim \mathcal{GP}\left(\mu\left(\mathbf{x}\right), k\left(\mathbf{x}, \mathbf{x}^\prime\right)\right)$,
where, $\mu(\cdot)$ and $k(\cdot)$ are the mean and kernel functions respectively, which completely describe our process \cite{rasmussen2006gaussian}. As is common, we assume a zero-mean prior throughout this work, using the kernel as the primary means of modeling the variation in our function over its domain. A popular choice for GP kernels is the Mat\'ern class of covariance function, which takes the form,
$k_\nu(\mathbf{x}, \mathbf{x}^\prime) = \sigma^2 \frac{2^{1-\nu}}{\Gamma(\nu)} \left(\frac{r\sqrt{2\nu}}{\kappa} \right)^\nu K_\nu \left(\frac{r\sqrt{2\nu}}{\kappa} \right)$,
where $r=\lVert \mathbf{x}-\mathbf{x}^\prime \rVert$ and $K_\nu$ is a modified Bessel function. We define $\boldsymbol{\theta} = \{ \sigma^2, \kappa, \nu\}$ to be the set of kernel hyperparameters; $\sigma^2$ controls the variance of the GP, $\kappa$ the lengthscale of its variation and $\nu$ its degree of differentiability.

\subsubsection{Inference:} Using Bayes' Rule, we can condition our GP on the training data and derive closed form expressions for the posterior mean and covariance:
\begin{align}
\label{eq:post_mu}
\boldsymbol{\mu}_{\text{post}} &= \mathbf{K}_{*}(\mathbf{K} + \sigma_y^2 \mathbf{I})^{-1} \mathbf{y}, \\
\label{eq:post_sigma}
\boldsymbol{\Sigma}_{\text{post}} &= \mathbf{K}_{**} - \mathbf{K}_* (\mathbf{K} + \sigma_y^2 \mathbf{I})^{-1} \mathbf{K}_*^\top .
\end{align}
Generally, the noise variance $\sigma_y^2$ and the kernel function hyperparameters $\boldsymbol{\theta}$ are optimised via maximisation of the log-marginal likelihood, which can also be derived analytically. Where $\mathbf{X} \in \mathbb{R}^{N \times P}$ and $\mathbf{y} \in \mathbb{R}^{N}$ are matrix and vectorial representations of our training inputs and targets respectively, the log-marginal likelihood takes the form \cite{rasmussen2006gaussian}, 
\begin{align} \label{eq:ll}
\begin{split}
    \log p(\mathbf{y} \mid \mathbf{X}, \boldsymbol{\theta}, & \sigma_y^2) = -\frac{1}{2} \mathbf{y}^\top \left(\mathbf{K} + \sigma_y^2\mathbf{I} \right)^{-1} \\ &- \frac{1}{2} \log \mid \mathbf{K} + \sigma_y^2\mathbf{I} \mid - \frac{N}{2} \log (2\pi).
\end{split}
\end{align}

\subsubsection{Kernels on manifolds:} Many different kernel functions for GPs exist, and choosing a kernel is in itself a model selection problem as some kernels are more suited to modeling certain types of data. However, one characteristic which many kernels share is that they are defined using Euclidean distance. This presents an issue should we wish to use a GP to model variation in a quantity over a non-Euclidean space. Borovitskiy \textit{et al.} \cite{borovitskiy2020matern} proposed a solution to this problem in the form of an extension to the Mat\'ern kernel, which allows for modeling of functions whose domains are compact Riemannian manifolds. The approach proposed by the authors involves two stages. Firstly, numerical estimation of the eigenvalues $\lambda_n$ and eigenfunctions $f_n$ corresponding to the Laplace-Beltrami operator of the given manifold is performed. Secondly, for a manifold of dimensionality $d$, the kernel is approximated using a finite truncation of:
\begin{equation} \label{eq:kernel}
    k_{\nu}(\mathbf{x}, \mathbf{x}^\prime) = \frac{\sigma^2}{C_{\nu}} \sum_{n=0}^{\infty} \left(\frac{2\nu}{\kappa^2} + \lambda_n \right)^{-\nu - \frac{d}{2}} f_n(\mathbf{x}) f_n(\mathbf{x}^\prime) ,
\end{equation}
where, $C_{\nu}$ is a normalizing constant. The hyperparameters $\sigma^2$, $\kappa$ and $\nu$ have similar interpretations to those introduced for the conventional Euclidean Mat\'ern kernel.

\subsection{Greedy subset-of-data algorithm}
\label{sec:sod}
A major challenge which arises when working with GPs in practice is the $\mathcal{O}(N^3)$ complexity associated with performing exact inference, which arises due to the matrix inversions in Equations \eqref{eq:post_mu} and \eqref{eq:post_sigma}. To circumvent this issue, numerous formulations of \textit{sparse GPs} have been proposed, many of which are based on approximate inference techniques and concepts such as \textit{inducing points} \cite{liu2020gaussian}. In this work however, we consider the \textit{subset-of-data} (SoD) approach. 
As explained in Section 8.3.3 of \cite{rasmussen2006gaussian}, it is a conceptually simple form of sparse approximation which allows for exact Bayesian inference. In this setting, rather than modifying the formulation of the GP itself, we simply perform exact inference using a carefully selected subset of $M (<< N)$ observations. Specifically, for our case we modify the greedy SoD approach of \cite{lalchand2018fast}, which uses a selection criterion to sequentially construct a subset of size $M$ which is representative of our full training set of $N$ observations. We use this technique for GP sparsification in order to construct a set of inducing points for a point cloud which are best capable of representing the changes in surface variation over the cloud; this set of points forms our simplified point cloud. The original method involves randomly selecting one initial inducing point and then adding one point to the set at each iteration, however we have employed \textit{farthest point sampling} (FPS) for selecting a set of initial inducing points instead of one, and we add several points to our set of inducing points at each iteration. Our approach is explained in further detail in Section \ref{sec:methodology}.

Our method forms a simplified point cloud which is a subset of the original, thus the optimization problem is a discrete one. There has been recent work on inducing point optimization on discrete domains \cite{fortuin2021sparse}, however such methods only obtain comparable performance to methods based on greedy selection of the inducing points from the input domain, which are considerably conceptually simpler. The main disadvantage of a greedy approach is that the training set does not necessarily span the whole input domain, however in our setting this is indeed the case, making our application especially well-suited to a greedy approach. Additionally, our proposed method allows us to obtain competitive results for clouds containing millions of points, whilst still employing exact Bayesian inference rather than approximate variational schemes, which can often underestimate the variance of the posterior distribution \cite{blei2017variational}.

\section{Point cloud simplification with Riemannian Gaussian processes} 
\label{sec:methodology}
In this section, we outline our GP-based approach with the help of a concise algorithm. We can represent a point cloud of size $N$ as a set of 3D Euclidean coordinates $P = \{\mathbf{x}_i\}^N_{i=1}$, where $\mathbf{x}_i \in \mathbb{R}^{3}$. The surface variation $y_i \in \mathbb{R}$ at each point in $P$ can be computed using Equation \eqref{sf}. Using this data we formulate a regression problem, whereby we employ a GP with a Mat\'ern kernel defined on a Riemannian manifold (as described in Section \ref{sec:riemannian_gp}) to predict the surface variation from the coordinates of each point. We then employ the greedy subset-of-data scheme discussed in Section \ref{sec:sod} in order to obtain a simplified set of $M (<< N)$ 3D coordinates, $P_{\text{simp}} = \{\mathbf{x}_j\}_{j=1}^M$, where $P_{\text{simp}} \subset P$.

We formally outline our proposed approach in Algorithm \ref{alg:simpli}.
$\text{FPS}(P, k_\text{init})$ denotes a function which selects $k_\text{init}$ initial points from $P$ using FPS; we use this to initialise our active set $P_{\text{simp}}$ with an initial set of points from across the point cloud. $\text{MAX}(\mathbf{s}, R, k_\text{add})$ selects the points from the remainder set $R$ which are associated with the $k_\text{add}$ largest values in our selection criterion vector $\mathbf{s}$. The notation $\mathbf{y}(R)$ denotes a vector containing the target surface variation values associated with each of the points contained within the set $R$. At each step $t$ of the algorithm, we update the posterior mean $\boldsymbol{\mu}_t$ and covariance $\boldsymbol{\Sigma}_t$ using Equations \eqref{eq:post_mu} and \eqref{eq:post_sigma} respectively, where the active set $P_{\text{simp}}$ is used as training data, whilst the remainder set $R$ is unseen test data.

\begin{algorithm}
\caption{GP-based simplification algorithm}\label{alg:simpli}
\begin{algorithmic}
\State \textbf{Data:} $P$, $\mathbf{y}$, $M$, $k_\text{init}$, $k_\text{add}$,  $k_\text{opt}$, GP prior $\mathcal{GP}(0, k(\cdot, \cdot))$, where $k$ is defined in Eq. \eqref{eq:kernel}
\State \textbf{Result:} $P_{\text{simp}}$
\State $P_{\text{opt}} \leftarrow$ random subset of  $k_\text{opt}$ points from $P$;
\State Optimise GP hyperparameters using Eq. \eqref{eq:ll}, $P_{\text{opt}}$ and $\mathbf{y}(P_{\text{opt}})$;
\State Active set $P_{\text{simp}} \leftarrow \text{FPS}(P, k_\text{init})$;

\State Remainder set $R \leftarrow P - P_{\text{simp}}$;
\While{$\lvert P_{\text{simp}} \rvert < M$}
\State Compute $\boldsymbol{\mu}_t$ and $\boldsymbol{\Sigma}_t$ using Eq. \eqref{eq:post_mu} and \eqref{eq:post_sigma};
    
\State $\mathbf{s} \leftarrow \sqrt{\text{diag}(\boldsymbol{\Sigma}_t)}+\lvert \boldsymbol{\mu}_t - \mathbf{y}(R) \rvert$;

\State $P_{\text{simp}} \leftarrow P_{\text{simp}} + \text{MAX}(\mathbf{s}, R, k_\text{add})$;
\State $R \leftarrow R - \text{MAX}(\mathbf{s}, R, k_\text{add})$;
\EndWhile
\end{algorithmic}
\end{algorithm}
To clarify, we predict the surface variation and the uncertainty values for $R$ based on $P_\text{simp}$ at each iteration of our algorithm. The selection criterion which we use favours selection of points within the original cloud which lie in regions of high predictive uncertainty and/or error. By selecting a set of points using this criterion, we form a simplified cloud which implicitly favours selection of points surrounding finer details within the cloud, where the error and uncertainty is likely to be high if we have not yet selected a sufficient number of points around said location.

As $P_\text{simp}$ grows with each iteration to be gradually more representative of our input data, the uncertainty and predicted surface variation values for points in $R$ also change. For example, consider two neighbouring points on the tip of one of the Stanford bunny’s ears, and assume that neither of them are currently in $P_\text{simp}$. If one of these points is added to $P_\text{simp}$, the elements of the uncertainty $\sqrt{diag(\boldsymbol{\Sigma_t})}$ and error $|\boldsymbol{\mu}_t - \mathbf{y}(R)|$ associated with the second point will decrease, and in subsequent iterations it may no longer be one of the top-ranked points based on the selection metric $\mathbf{s}$.

\section{Empirical evaluation}
\label{sec:experiments}

In this section, we extensively evaluate the proposed simplification method using various point cloud datasets and processing techniques. First, we compare our simplification technique both quantitively and qualitatively using benchmark object level point clouds as given in subsection \ref{sec:opc}. Second, in subsection \ref{sec:reg} we extend the use-case of our algorithm as a time and memory efficient pre-processing step for the downstream task of point cloud registration. Moreover, we provide some experiments on scene level and self-acquired point clouds along with ablation studies in the supplementary material (Section 2).
%

\subsection{Benchmark object level point clouds}
\label{sec:opc}

\subsubsection{Evaluation criteria:}
\label{sec:eval_criteria}
In order to evaluate the performance of our method in comparison to other simplification techniques, we firstly use each simplified point cloud obtained from three object level point clouds to form simplified meshes, using \textit{screened Poisson surface reconstruction} \cite{kazhdan2013screened}. We can then compute the reconstruction errors between the original meshes, and the reconstructed meshes formed from our simplified clouds. Specifically, we choose to evaluate the mean and maximum \textit{Hausdorff distance} \cite{cignoni1998metro}. Evaluating the error associated with mesh reconstruction is effective at quantifying the ability of each method to preserve features from the original cloud, as accurate reconstruction of a mesh from a simplified point cloud requires that a high density of points be placed in the vicinity of finer details within the cloud. The \textit{MeshLab} software 
was used to reconstruct all surfaces and compute the Hausdorff distances. Also, given that one of our primary aims is to preserve sharp features within each point cloud, we also report the \textit{average surface variation} over each simplified point cloud. The surface variation at each point is computed using the approach described in Section \ref{sec:surface}.

\subsubsection{Baselines:}
We use the aforementioned evaluation procedure to compare our method (denoted \textit{GP}) empirically to a number of competing simplification techniques discussed in Section \ref{sec:rw}. We compare our approach to \textit{PC-Simp}, \textit{AIVS}, \textit{Potamias et al.}, \textit{HC} and \textit{WLOP}, with the latter two approaches implemented using the \textit{CGAL} library.  Additionally, we provide a visual comparison of our simplification method with \textit{APES}.
For the HC method, the size and variation parameters discussed in Section \ref{sec:rw} were manually tuned to obtain approximate desired simplified sizes. Also, as noted in Section \ref{sec:rw}, we use the non-curvature aware version of the AIVS algorithm, as there is no available open-source implementation of the curvature-aware variant.

\begin{figure}
\centering
  \includegraphics[scale=0.37, trim=0cm 4cm 0cm 4.8cm, clip]{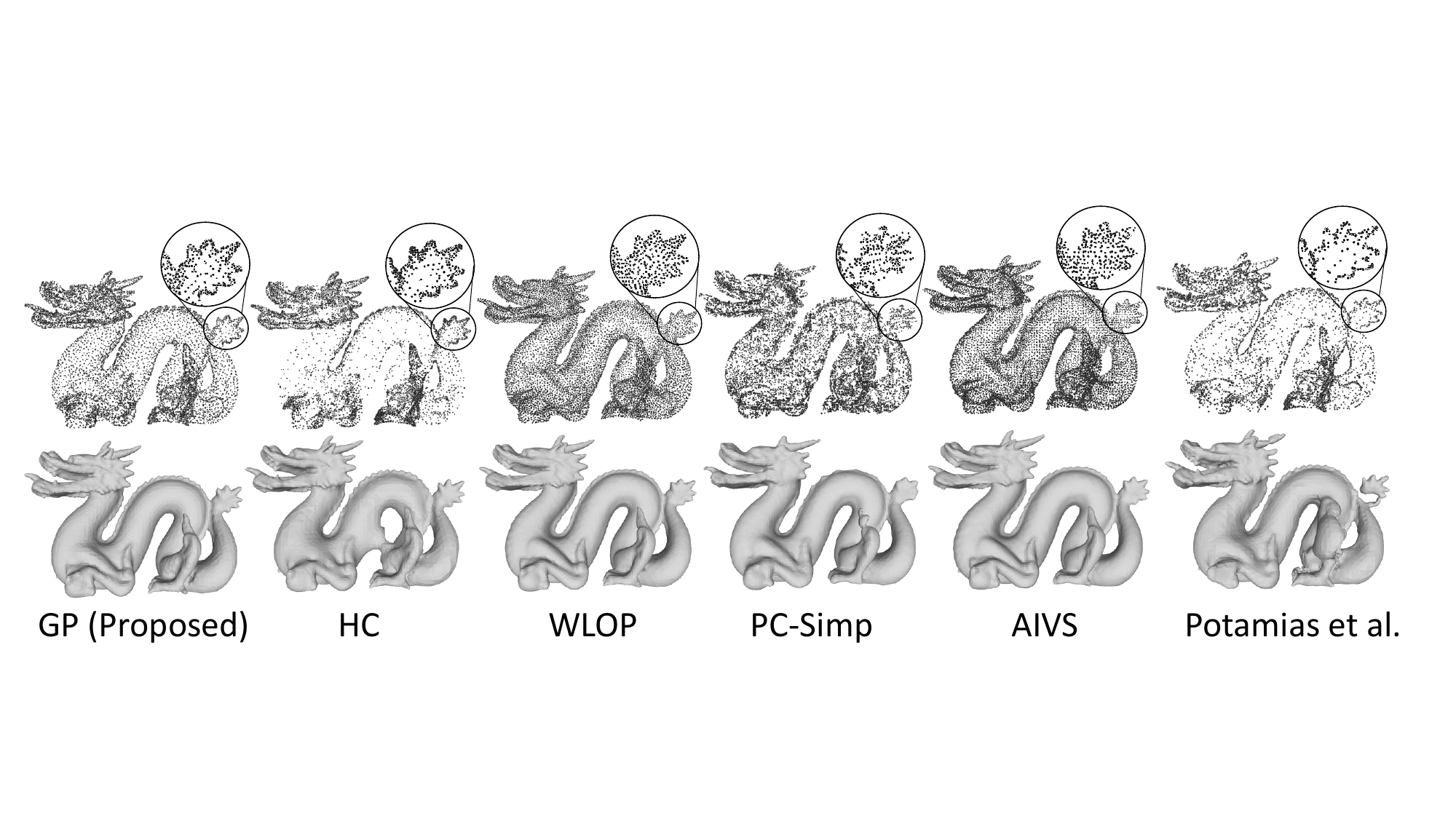}
           \caption{Simplified representations of the Dragon point cloud for simplification ratio $\alpha=0.03$ (top row) and associated reconstructed meshes (bottom row) for all evaluated simplification techniques.}
  \label{fig:dragon_all}

\end{figure}

\subsubsection{Experimental details:}
We evaluate our proposed method and the aforementioned baselines on three complex object-level point clouds from the Stanford 3D Scanning Repository \cite{levoy2005stanford}, namely \textit{Armadillo} ($N = 1,72,974$), \textit{Dragon} ($N = 4,37,645$) and \textit{Lucy} ($N = 1,40,27,872$). Let the \textit{simplification ratio} be defined as $\alpha = M/N$. In this work we focus on the challenging regime where we wish to significantly reduce the size of the cloud, such that $\alpha << 1$. It is in this regime that feature-preserving techniques such as ours become particularly important, as we do not have a large number of points to select, thus we must efficiently select points which allow us to capture the salient features of the original cloud. We chose $\alpha$ for each cloud by finding the minimum $\alpha$ at which all evaluated techniques were capable of forming simplified clouds from which meshes visually comparable to the original meshes could be generated \cite{levoy2005stanford}. This value varies depending on the surface complexity of each cloud, thus for \textit{Armadillo}, \textit{Dragon} and \textit{Lucy} we chose $\alpha = 0.05$, $0.03$ and $0.002$ respectively. Additionally, we also visually evaluate the point cloud simplification results of all aforementioned techniques on a noisy Armadillo from the PCPNet dataset \cite{guerrero2018pcpnet}, with $\alpha=0.05$. This corresponds to the original Armadillo model surface sampled $10^5$ times ($N = 1,00,000$), with Gaussian noise (of standard deviation $\sigma = 2.5\times 10^{-3} \times d$, where $d$ is bounding box diagonal length) added to every point position. We also perform the same evaluation for three objects, an airplane, a glass and a toilet ($N = 2,048$ for all three) from ModelNet40 dataset \cite{wu20153d} to compare our method with APES.

\begin{figure}[h]
\centering
    \includegraphics[scale=0.37, trim=0.3cm 10cm 0.3cm 0.1cm]{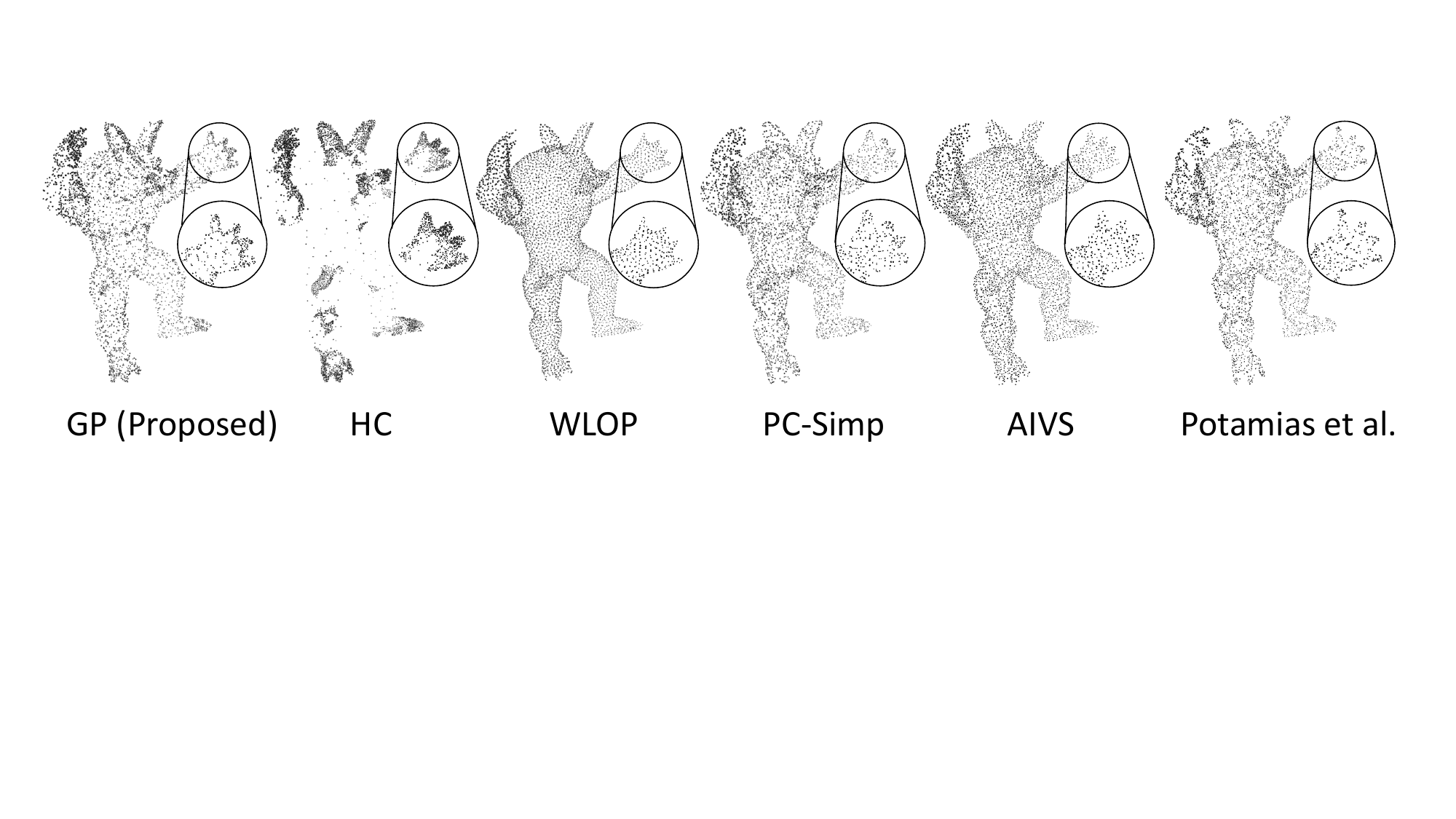}
           \caption{Simplification results of a noisy Armadillo with Gaussian noise added to every point position (of standard deviation $\sigma = 2.5\times 10^{-3} \times d$, where $d$ is the bounding box diagonal length) for simplification ratio $\alpha=0.05$ for all evaluated simplification techniques.} 
           \vspace{-0.3cm}
  \label{fig:noisy}
\end{figure}

\begin{figure}[h]
\centering
    \includegraphics[scale=0.56, trim=4cm 0cm 7cm 0cm]{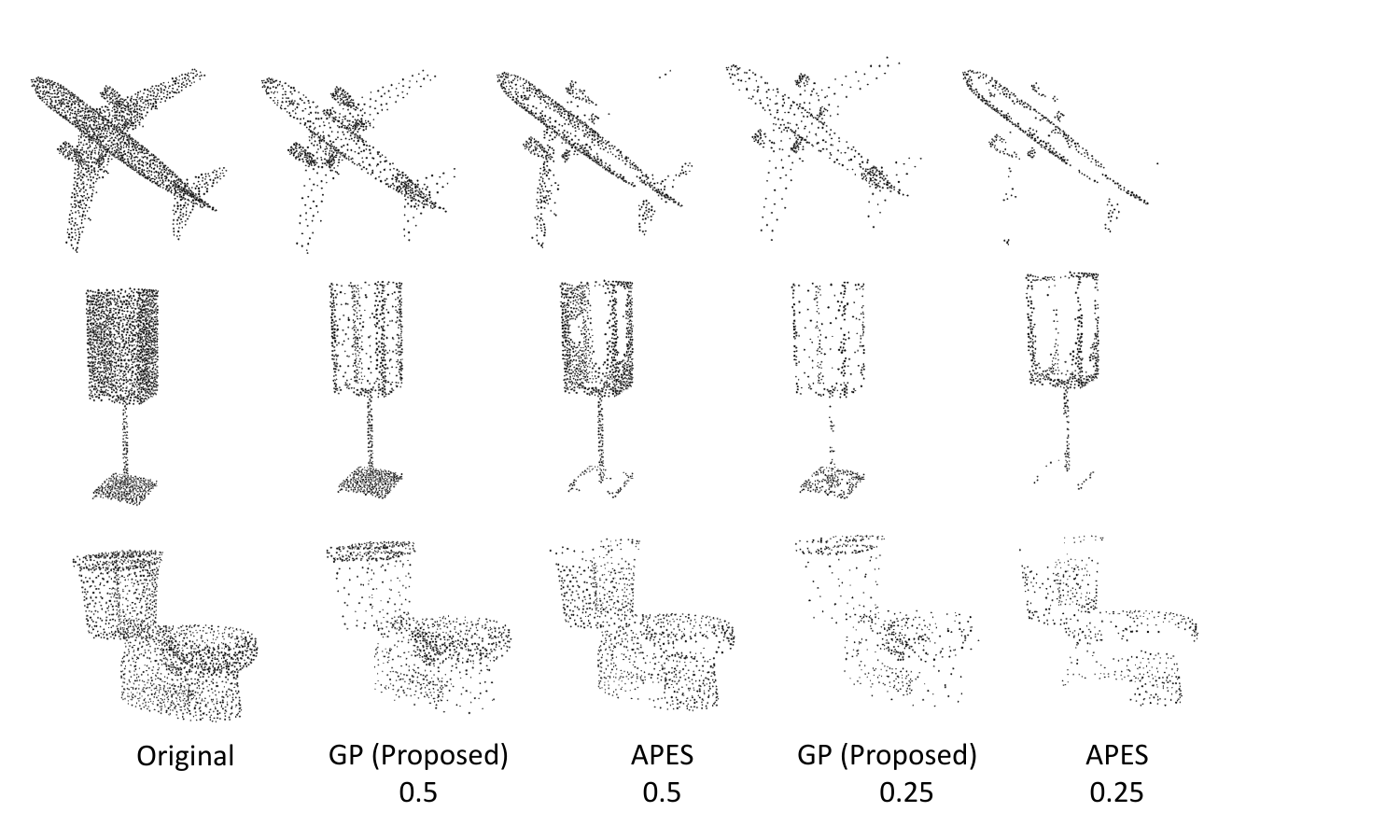}
           \caption{Simplification results of an airplane, a glass and a toilet point cloud for simplification ratios $\alpha=0.25$ and $0.5$ using APES and GP-based simplification.} 
           \vspace{0cm}
  \label{fig:apes}
\end{figure}

\subsubsection{Discussion:}
From the results presented in Table \ref{tab:results}, it is clear that our proposed method is capable of comparable empirical performance to many of the existing methods for simplifying point clouds. The GP-based approach outperforms the AIVS baseline across all experiments and metrics, and outperforms the PC-Simp baseline on all but the mean Hausdorff distance for the Armadillo experiment. Moreover, our algorithm also runs considerably faster than both of these approaches. Note that due to the scale of \textit{Lucy}, we were unable to evaluate PC-Simp on this cloud as it was taking more than two hours to run. 

\begin{table*}[h] \vspace{-0.4cm}
    \caption{Empirical results and total runtimes (time taken by surface variation computation and simplification) for all tested simplification methods and point clouds. We report the maximum and mean Hausdorff distances between the original meshes, and the meshes reconstructed from the simplified point clouds. Also reported is the average surface variation over each simplified point cloud. Best, second-best and third-best results are in \textcolor{red}{red}, \textcolor{ForestGreen}{green} and \textcolor{blue}{blue} respectively. It is worth mentioning that as per the evaluated metrics, our algorithm mostly stays within the top three methods.}
    \centering
    \resizebox{\linewidth}{!}{
    \begin{tabular}{ccccccccccccc}
    \toprule
    {} & \multicolumn{3}{c}{Mean Hausdorff Distance ($\downarrow$)} & \multicolumn{3}{c}{Max. Hausdorff Distance ($\downarrow$)} & \multicolumn{3}{c}{Mean Surface Variation ($\uparrow$) }  & \multicolumn{3}{c}{Total Time (s) ($\downarrow$) } \\
    \cmidrule(lr){2-4} \cmidrule(lr){5-7} \cmidrule(lr){8-10} \cmidrule(lr){11-13}
    {} & Armadillo & Dragon & Lucy & Armadillo & Dragon & Lucy & Armadillo & Dragon & Lucy & Armadillo & Dragon & Lucy\\
    \midrule
     GP (ours) & 0.246 & \textcolor{ForestGreen}{0.000246} & \textcolor{red}{1.11} & \textcolor{red}{3.26} & \textcolor{ForestGreen}{0.00457} & 195.78 & \textcolor{ForestGreen}{0.0728} & \textcolor{ForestGreen}{0.0546} & \textcolor{ForestGreen}{0.0724} & \textcolor{blue}{0.8} & \textcolor{blue}{1.4} & \textcolor{blue}{12.9} \\
     HC & 0.374 & 0.000758 & \textcolor{ForestGreen}{1.14} & \textcolor{red}{3.26} & 0.0141 & \textcolor{ForestGreen}{195.41} & \textcolor{red}{0.0803} & \textcolor{red}{0.0686} & \textcolor{red}{0.0762} & \textcolor{ForestGreen}{0.1} & \textcolor{ForestGreen}{1.1} & \textcolor{ForestGreen}{10.0} \\
     WLOP & \textcolor{red}{0.197} & \textcolor{red}{0.000188} & \textcolor{blue}{1.29} & 4.14 & \textcolor{red}{0.00417} & \textcolor{blue}{195.52} & \textcolor{blue}{0.0557} & 0.0413 & 0.0631 & 3.5 & 6.5 & 84.2 \\
     PC-Simp & \textcolor{blue}{0.241} & 0.000487 & - & 5.48 & 0.00802 & - & 0.0364 & 0.0433 & - & 132.6 & 245.5 & - \\
     AIVS & 0.715 & 0.000638 & 8.75 & \textcolor{blue}{4.11} & \textcolor{blue}{0.00539} & 196.45 & 0.0513 & \textcolor{blue}{0.0441} & \textcolor{blue}{0.0666} & 17.2 & 44.6 & 1983.5  \\
     Potamias et al. & \textcolor{ForestGreen}{0.215} & \textcolor{blue}{0.000599} & 4.28 & \textcolor{ForestGreen}{3.47} & 0.00933 & \textcolor{red}{190.00} & 0.0478 & 0.0650 & 0.0511 & 	
\textcolor{red}{0.00060} & \textcolor{red}{0.00070} & \textcolor{red}{0.00212}  \\
    \bottomrule
    \\
    \end{tabular}
    }
    \label{tab:results}
    \vspace{-0.5cm}
\end{table*}

HC and Potamias \textit{et al.} are the only baselines with shorter runtimes than our method, and obtain maximum Hausdorff distances comparable to those obtained by our approach. However, as discussed in Section \ref{sec:rw}, tuning the user-specified HC parameters make striking a balance between feature preservation and retaining a sufficient density of points across the cloud relatively challenging. Moreover, there is no control over the size of the simplified cloud, as discussed by the authors \cite{pauly2002efficient} and in subsequent work \cite{lv2021approximate}. We tuned this baseline to attempt to balance this trade-off, and whilst the HC-simplified clouds shown in Figures \ref{fig:dragon_all} and \ref{fig:noisy} here, and Figure 3 of the supplementary material, do have clearly preserved features (an observation supported by the high mean surface variation across all clouds), the density of points away from these areas is very low. This leads to inferior mesh reconstructions compared to our approach, as evidenced by the fact that we obtain superior mean Hausdorff distance compared to HC across all three clouds. 

Since results and inference times for the Potamias \textit{et al.} approach were provided by the author of the paper, we do not have knowledge of the exact details of their experimental setup, especially the time required in hours to train the model. As mentioned in Section \ref{sec:rw}, their learning-based approach demands huge datasets to train on, which not only increases the computational requirements but also limits their method's generalizability. When compared to our method quantitatively, our method generally gives superior results, except for two of the nine error metric values. This is supported by the quality of their simplified point clouds and the corresponding reconstructions shown in Figures \ref{fig:dragon_all} and \ref{fig:noisy} here, and in Figures 2, 3 and 5 of the supplementary material. Although their method performs best in the case of Lucy's maximum Hausdorff distance, in reality their simplified cloud gives arguably the poorest qualitative reconstruction result amongst all of the other baselines. As expected, the inference time of their approach is the lowest of all the baselines, because of their neural network-based approach, which involves pre-training.

The WLOP baseline does not efficiently preserve the features and favours uniformly covering the domain of the original cloud. Therefore, the mean surface variation of the WLOP simplified clouds is lower, but overall the Hausdorff distances obtained from the reconstructed meshes are superior to those obtained by our method. However, it is noteworthy that on the largest and unarguably the most challenging point cloud, \textit{Lucy}, our method achieves a superior mean Hausdorff distance as compared to all of the other techniques evaluated, including WLOP. Additionally, WLOP is significantly slower than our approach, as shown in Table \ref{tab:results}. Our surface variation computation is currently performed on a CPU, therefore further improvements to the runtimes of our method shown could be achieved by re-implementing this in a GPU-compatible framework.

Overall, these results show that our approach provides a computationally efficient option for performing point cloud simplification in settings where the user wishes to strike a balance between preserving high fidelity around sharp features in the cloud, and ensuring that the simplified cloud covers the manifold defined by the original cloud with a sufficient density of points. This is important for generating reconstructions which resemble the original meshes, as is evident from visual inspection of the reconstruction results in Figure \ref{fig:dragon_all} here and Figure 3 of the supplementary material. In terms of surface reconstruction, our method clearly outperforms all of the other techniques for the \textit{Dragon} (compare the tail, teeth, horns and the face detailing for all methods and additionally the curved body for HC) and the \textit{Armadillo} (compare the ears, hands and feet across all the methods) and gives competitive results for \textit{Lucy}, shown in Figure 2 of the supplementary material. We highlight once again the poor surface reconstructions resulting from the Potamias \textit{et al.} simplified clouds, compared to those obtained using all of the other baselines. Again, visual inspection of the simplification results for the noisy Armadillo in Figure \ref{fig:noisy} demonstrates the balanced feature-sensitivity of our method in comparison to others. We experiment with more noise levels in the supplementary material (Section 2). Finally, from Figure \ref{fig:apes} we can see how the edge-sampling-based APES simplified clouds have several missing portions including object edges, whereas our method enhances the salient features and captures the overall object structure simultaneously. We do not provide corresponding surface reconstructions and hence quantitative results for this baseline because their low simplified point cloud sizes ($N = 1,024$ and $512$) and aforementioned missing areas will always result in open meshes.


The $\mathcal{O}(M^3)$ and $\mathcal{O}(M^2 N)$ complexities associated with training and prediction respectively in the greedy inference scheme described in Section \ref{sec:sod} allow for increased scalability compared to typical GP regression, in which inference has $\mathcal{O}(N^3)$ complexity. The scalability of our approach is limited by the fact that, as in a conventional exact GP, we have a storage demand associated with $\mathbf{K}$ matrix which scales according to $\mathcal{O}(N^2)$. However, we can circumvent this issue when $N$ is very large by simply using Algorithm \ref{alg:simpli} with a randomly selected subset of $P$. For \textit{Armadillo} and \textit{Dragon} we obtain the above results with just 25,000 randomly selected points. For a large point cloud such as \textit{Lucy}, we obtain competitive results using a subset of just 40,000 points to run our simplification algorithm.

\subsection{Point cloud registration}
\label{sec:reg}
As discussed earlier, PC simplification has benefits for many downstream tasks, not solely surface reconstruction. In Table \ref{tab:register} we present registration results on some simplified clouds. We firstly translate and rotate the original, HC and GP-simplified clouds in the same fashion, before performing global and ICP point-to-point registration \cite{besl1992method} with the \textit{Open3D} package \cite{Zhou2018}; visualisations are available in the supplementary material (Figure 4). Our GP-simplified cloud allows for quicker registration and leads to superior inlier RMSE.
\begin{table}
    \caption{Inlier RMSE and time taken for global and ICP registration. Best results are in \textcolor{red}{red}, whilst second-best are in \textcolor{ForestGreen}{green}.} 
    \centering
    \begin{tabular}{ccc@{\hskip 0.5cm}ccc}
    \toprule
    {} & \multicolumn{2}{c}{Inlier RMSE ($\downarrow$)} & \multicolumn{3}{c}{Time (s) ($\downarrow$)}  \\
    \cmidrule(lr){2-3} \cmidrule(lr){4-6} 
    {} & Global ($10^{-3}$) & ICP ($10^{-7}$) & Global & ICP & Total \\
    \midrule
     Original & \textcolor{ForestGreen}{4.76} & \textcolor{red}{4.08} & \textcolor{red}{0.017} & 1.448 & 1.465\\
     HC & 5.41 & \textcolor{red}{4.08} & \textcolor{ForestGreen}{0.018} & \textcolor{ForestGreen}{0.046} & \textcolor{ForestGreen}{0.064}\\
     GP (ours) & \textcolor{red}{3.91} & \textcolor{red}{4.08} & \textcolor{red}{0.017} & \textcolor{red}{0.040} & \textcolor{red}{0.057} \\
    \bottomrule
    \\
    \end{tabular}

    \label{tab:register}
\end{table}

\setlength{\tabcolsep}{4pt}

\vspace{-1cm}

\section{Conclusion}
\label{sec:conc}
In this work we have presented a novel, one-shot point cloud simplification algorithm capable of preserving both the salient features and the overall structure of the original point cloud. We reduce the cloud size by up to three orders of magnitude without the need for computationally intensive training on huge datasets. This is achieved via a greedy algorithm which iteratively selects points based on a selection criterion determined by modeling the surface variation over the original point cloud using Gaussian processes with kernels which operate on Riemannian manifolds. We show that our technique achieves competitive results and runtimes when compared to a number of relevant methods, outperforming all baselines tested in terms of mean Hausdorff distance on \textit{Lucy}, the largest and most complex point cloud we consider, consisting of approximately 14 million points. Our method can also be used to improve the computational efficiency of downstream tasks such as point cloud registration with no negative effects on the empirical performance.

\subsubsection{Future work:} 
Whilst Hausdorff distance is a useful metric, it is not the ideal candidate for assessing the feature sensitivity of a simplification algorithm, as it tends to return lower errors for more evenly distributed clouds. Whilst out of the scope of this work, there is a clear need for a well-defined and widely adopted error metric for curvature-sensitive simplification. Currently, the best way to evaluate this is a qualitative visual inspection of the resulting point cloud (or reconstructed mesh). This view is supported by the fact that some recent works employ user studies to evaluate their feature-preserving approaches \cite{potamias2022revisiting}.

In this work we study the setting where we enforce the restriction that the simplified cloud be a subset of the original; as discussed in Section \ref{sec:sod}, a greedy inference scheme is appropriate in this setting. However, this assumption could be relaxed and sparse GPs can be used to perform continuous optimization of the inducing points across the point cloud \cite{hutchinson2021vector}. This would allow occluded as well as outlier-ridden extremely noisy point clouds, where the original observations do not necessarily lie on the true surface of the manifold, to be denoised and/or simplified.


%
%
%
\bibliographystyle{abbrv}
\bibliography{references}

\newpage
\setcounter{section}{0}
\section*{Supplementary Material}
\section{Additional experimental details}
\label{sec:exp_det}
Our approach practically has no user-specified parameters, as it generates a simplified cloud based on the desired simplification ratio ($\alpha$). However, we supply to our algorithm (Algorithm 1 in main text) some fixed default values which work well for most point clouds. We set $k_{\text{opt}} = 200$ for all experiments as even with a small subset of the original point cloud, the GP typically converges on an optimal set of hyperparameters within 100 iterations. $k_{\text{init}}$ was chosen to be $1/3$ of the target number of points in the simplified cloud, a ratio which empirically works well across all point clouds tested. Finally, $k_{\text{add}}$ is determined adaptively based on $k_{\text{init}}$, $N$ and $M$. Our algorithm is implemented in the \textit{PyTorch} framework, 
and whilst the runtimes reported in Table 1 of main manuscript were achieved with GPU acceleration using an NVIDIA A100 with 80GB of RAM, our algorithm can also be run purely on a CPU. All baselines other than Potamias \textit{et al.} and APES were run on an Intel i7-11800H CPU with 32GB RAM.
\section{Additional visualisations and experiments}
\label{sec:add_vis}
On the following pages we present additional visualisations and quantitative comparisons to accompany the results presented in the main text. Figure \ref{fig:original} shows the some original meshes provided here for a better visual comparison. Figure \ref{fig:lucy_all} shows the surface reconstruction results on \textit{Lucy}. Figure \ref{fig:armadillo_all} shows the simplified clouds and reconstructed meshes for all techniques on the \textit{Armadillo} cloud. Figure \ref{fig:regis} is a qualitative comparison of the HC and GP-based approaches to performing point cloud registration on the \textit{Dragon} cloud.

Furthermore, we validate our technique's feature-sensitive approach on real-world scanning datasets captured using different acquisition devices. Firstly, we use a desk scene point cloud from the NYU Depth V2 dataset, derived from RGBD data acquired using RGB and Depth cameras from Microsoft Kinect. This cloud and the resulting simplification results from proposed and Potamias \textit{et al.} methods are shown in Figure \ref{fig:nyu}. Secondly, we acquired point clouds from three real-life objects: an angel figure, a human and a bike frame. They were captured using FARO's scan-in-a-box system, photogrammetry and AliceVision's Meshroom 
and Artec 3D's portable 3D scanner respectively. The three of them when simplified using our method with $\alpha= 0.04$, $0.05$ and $0.6$ respectively give results which are shown in Figure \ref{fig:new}. In line with the quantitative results provided in the main text, we also report the mean and maximum Hausdorff distances along with the mean surface variation values of all the aforementioned simplified point clouds in Table \ref{tab:results_extra}. As per the given evaluation metrics, this table yet again confirms the fact that our algorithm mostly stays within the top three simplification methods. Yet again, as mentioned in our conclusion, complete disagreement of quantitative and qualitative results for the NYU scene using Potamias \textit{et al.} and proposed method justifies our argument that visual inspection is the best way to evaluate the feature sensitivity of a method.

Finally, in Figure \ref{fig:angel} we show an extensive comparison of our method with all the mentioned simplification techniques for a range of simplification ratios for the angel point cloud. Again, our method works best for striking a trade-off between enhancing salient features and keeping the low curvature region well-sampled, and hence leads to the best reconstruction results ($\alpha=0.05$) when compared to the original angel mesh (Figure \ref{fig:original}). The reconstructions obtained from WLOP, PC-Simp, AIVS and Potamias \textit{et al.} simplified clouds are less detailed around salient features (compare the hair, fingers, face, wings and feet with the original angel mesh in Figure \ref{fig:original}) while the one from HC-simplified cloud is quite poor around the torso region of the angel because of less density of points around low curvature regions in the simplified cloud. 

\subsubsection{Ablation and noise studies: } 
\label{ablation_sec}
We conduct experiments to understand the significance of some parameters within our algorithm. These parameters were $k_{init}$, $i$, $k_{opt}$, and $k$ (the number of initial points chosen using FPS, the number of times the hyperparameters of the GP are optimized, the number of points used for hyperparameter optimization, and the number of neighbourhood points used for surface variation estimation). More specifically, while keeping all other parameters fixed and varying just one, we compute the Chamfer distance between the original and the simplified clouds of the Stanford bunny for the simplification ratio ($\alpha$) of 0.2. Moreover, we also quantify and extend the results of simplifying a noisy Armadillo in the main text by providing Chamfer distance between the original Armadillo with varying noise levels (Gaussian noise of standard deviation $n\times d$, where $d$ is the bounding box diagonal length, is added to every point position) and simplified versions of them for the simplification ratio ($\alpha$) of 0.04. All of these results are given in Table \ref{tab:ablation}.

\setcounter{figure}{0}
\setcounter{table}{0}

\begin{figure}[h]
     \centering
        \includegraphics[trim=2.2cm 6.5cm 5cm 2cm, clip, scale=0.7]{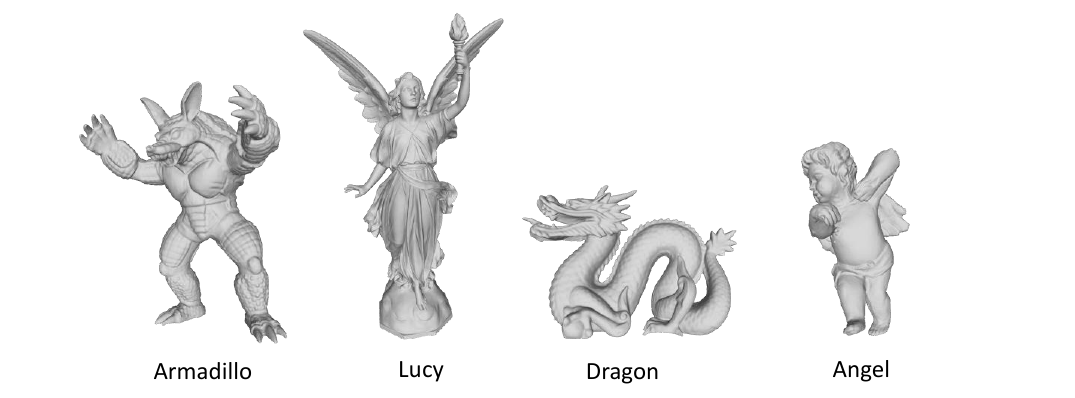}
         \caption{The original Stanford meshes and the angel mesh.}
         \label{fig:original}
\end{figure}

\begin{figure}
     \centering 
        \includegraphics[trim=0cm 3.5cm 0cm 3.7cm, clip, scale=0.35]{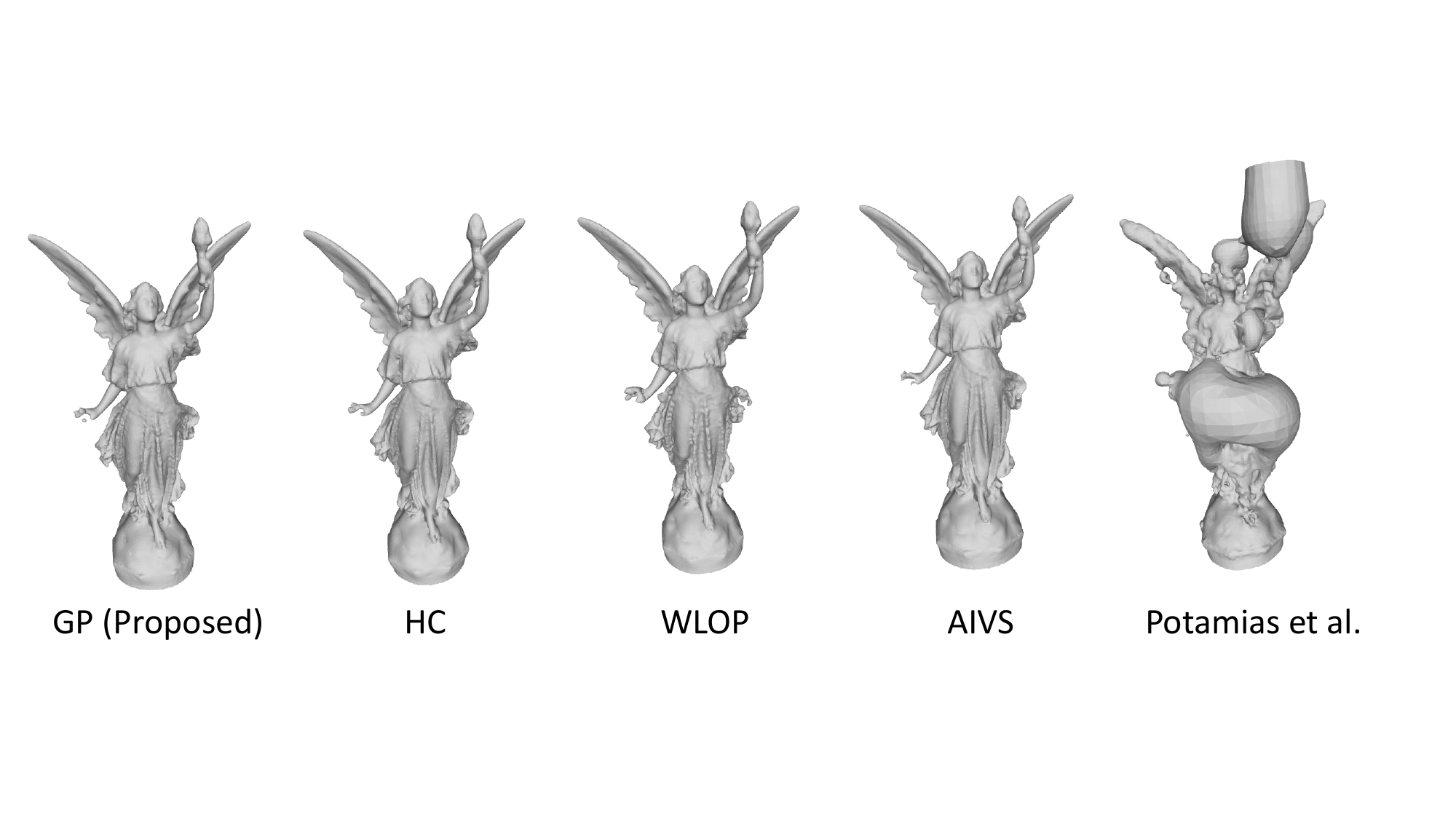}
         \caption{Surface reconstruction results of the simplified version of the Lucy point cloud for simplification ratio $\alpha=0.002$ for all evaluated simplification techniques except PC-Simp.}
         \label{fig:lucy_all}
\end{figure}
\begin{figure}
     \centering
        \includegraphics[trim=0cm 1.5cm 0cm 0cm, clip, scale=0.35]{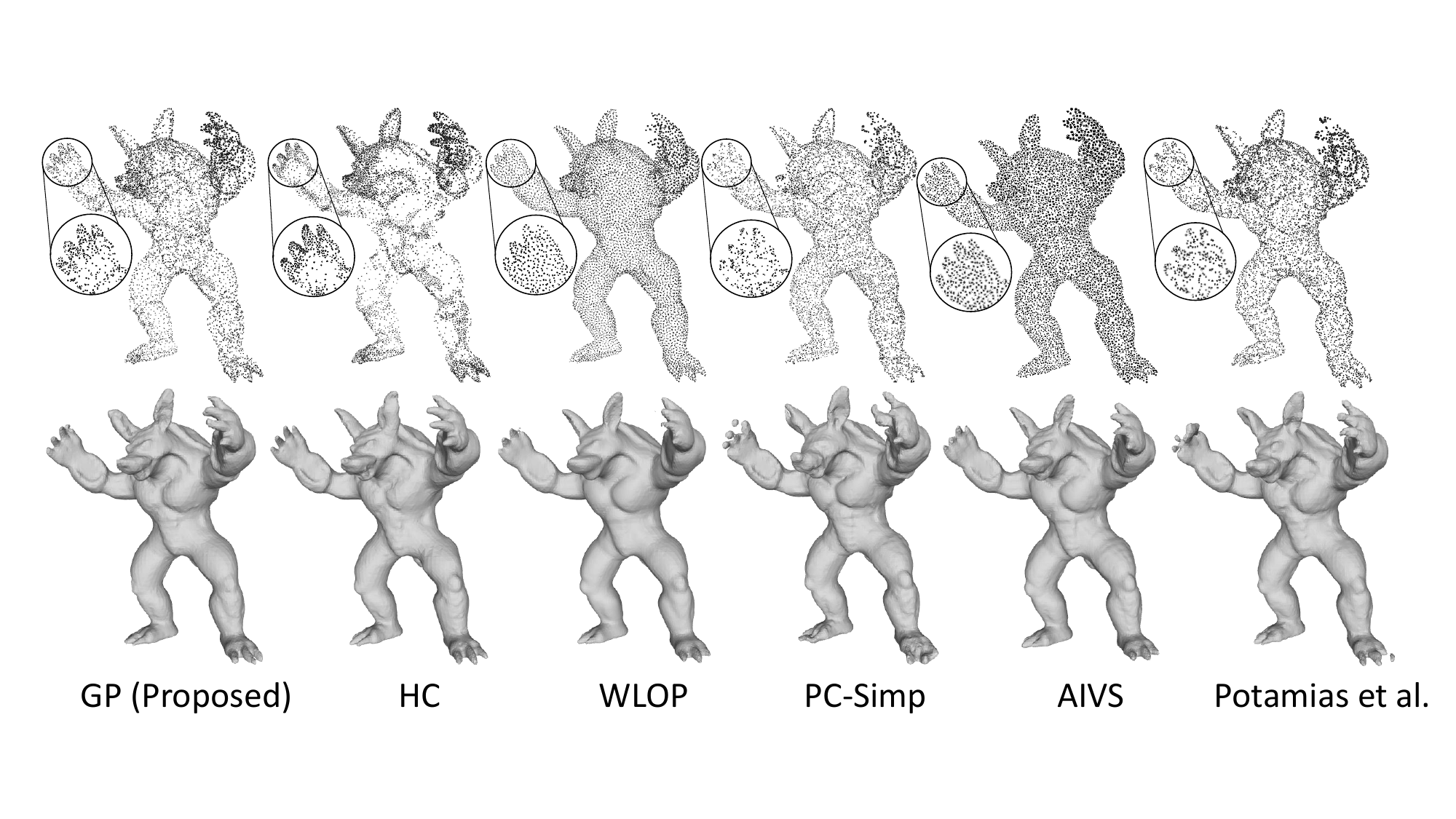}
         \caption{Simplified representations of the Armadillo point cloud for simplification ratio $\alpha=0.05$ (top row) and associated reconstructed meshes (bottom row) for all evaluated simplification techniques.}
    \label{fig:armadillo_all}
\end{figure}
%
%
\begin{figure}
     \centering
        \includegraphics[scale=0.35, trim=0.2cm 0cm 10cm 0cm, clip]{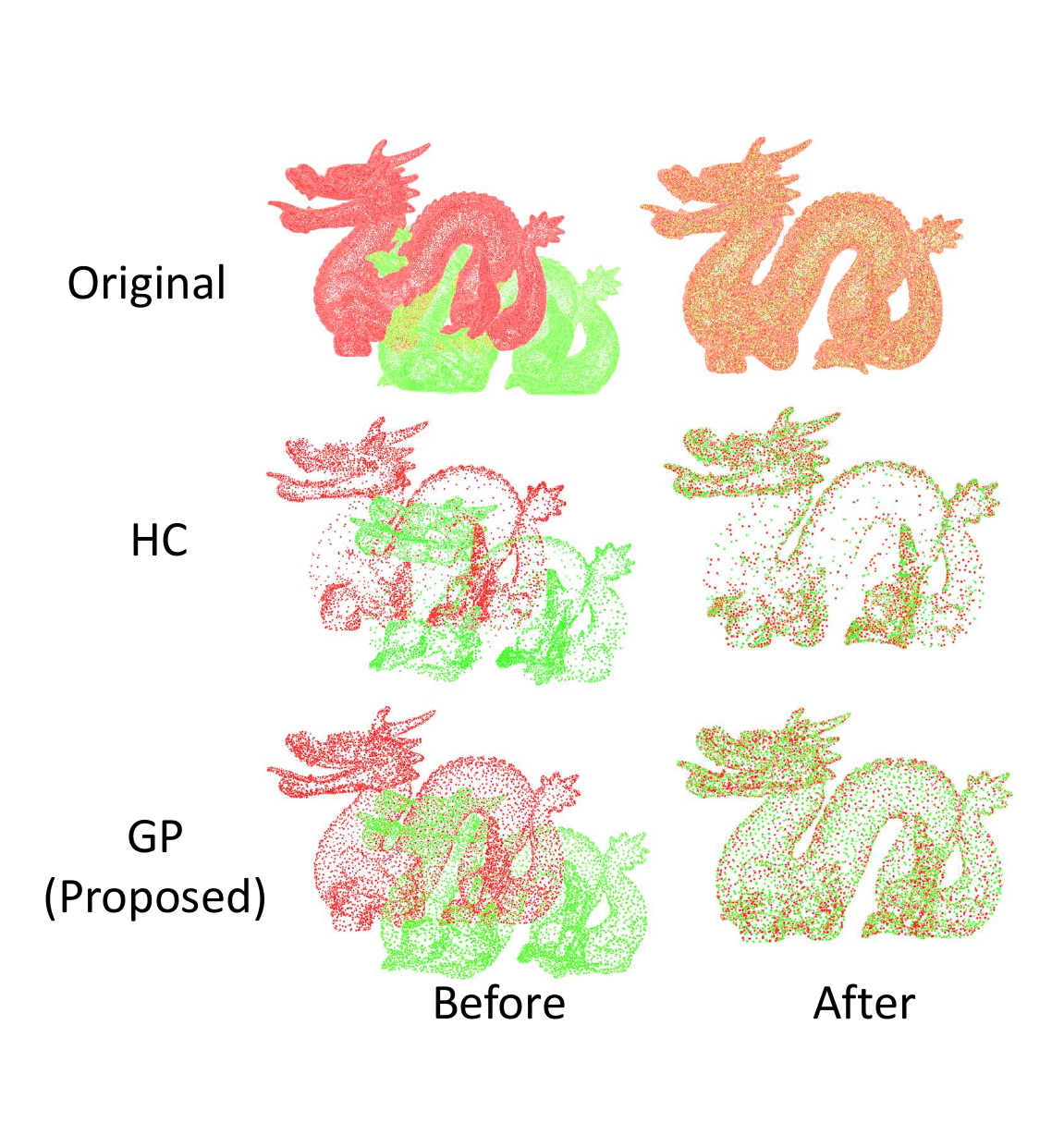}
         \caption{Global and ICP registration results shown for the original, HC and GP-simplified versions (simplification ratio $\alpha=0.03$) of the Stanford dragon.}
         \label{fig:regis}
\end{figure}



\begin{figure}
     \centering
        \includegraphics[trim=1cm 6cm 0cm 1.7cm, clip, scale=0.42]{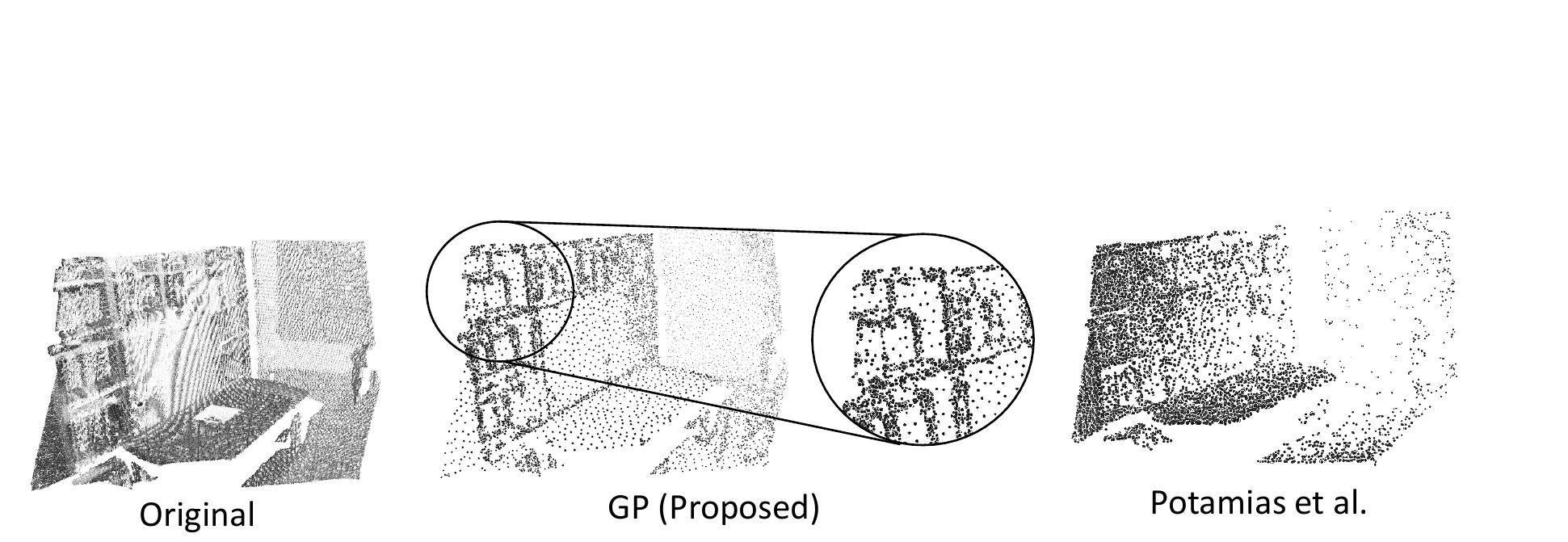}
           \caption{GP-based and Potamias \textit{et al.} simplification applied on a point cloud derived from real-life NYU Depth V2 Dataset's desk scene with simplification ratio $\alpha=0.03$.}
  \label{fig:nyu}
\end{figure}

\begin{figure}
     \centering \hspace*{-1cm}
        \includegraphics[trim=0cm 14cm 12cm 2.4cm, clip, scale=0.85]{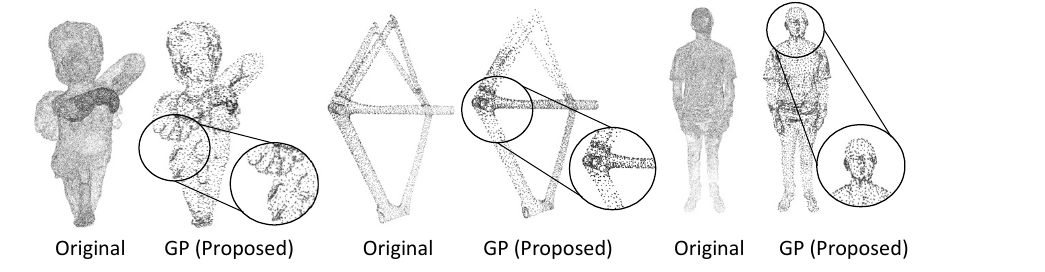} 
           \caption{Simplification of three self-acquired point clouds.}
  \label{fig:new}
\end{figure}

\vspace{5cm}
\begin{table*}
    \caption{Empirical results for all tested simplification methods for the remaining point clouds. We report the maximum and mean Hausdorff distances between the original meshes, and the meshes reconstructed from the simplified point clouds. Also reported is the average surface variation over each simplified point cloud. Best, second-best and third-best results are in \textcolor{red}{red}, \textcolor{ForestGreen}{green} and \textcolor{blue}{blue} respectively.}
    \centering
    \resizebox{\linewidth}{!}{
    \begin{tabular}{c@{\hskip 0.3cm}cccc@{\hskip 0.3cm}cccc@{\hskip 0.3cm}cccc}
    \toprule
    {} & \multicolumn{4}{c}{Mean Hausdorff Distance ($\downarrow$)} & \multicolumn{4}{c}{Max. Hausdorff Distance ($\downarrow$)} & \multicolumn{4}{c}{Mean Surface Variation ($\uparrow$) } \\
    \cmidrule(lr){2-5} \cmidrule(lr){6-9} \cmidrule(lr){10-13} 
    {} & Angel & Bike frame & Human & NYU & Angel & Bike frame & Human & NYU & Angel & Bike frame & Human & NYU\\
    \midrule
     GP (ours) & \textcolor{blue}{0.000172} & \textcolor{red}{0.001921} & \textcolor{ForestGreen}{0.196125} & \textcolor{blue}{0.045540} & \textcolor{ForestGreen}{0.001701} & \textcolor{blue}{0.019782} & \textcolor{red}{4.287749} & 1.704832 & \textcolor{ForestGreen}{0.0445} & \textcolor{blue}{0.1021} & \textcolor{blue}{0.0309} & 0.0487 \\
     HC & 0.000270 & \textcolor{blue}{0.001959} & {0.395263} & {0.048864} & 0.006358 & \textcolor{ForestGreen}{0.019743} & {10.380267} & \textcolor{ForestGreen}{1.483680} & \textcolor{red}{0.0509} & 0.1003 & \textcolor{red}{0.0407} & \textcolor{red}{0.0648} \\
     WLOP & \textcolor{red}{0.000098} & {0.006128} & \textcolor{red}{0.153166} & 0.046066 & \textcolor{red}{0.000920} & {0.086360} & \textcolor{ForestGreen}{4.410541} & \textcolor{blue}{1.671164} & 0.0294 & 0.0737 & 0.0169 & 0.0403 \\
     PC-Simp & \textcolor{ForestGreen}{0.000150} & \textcolor{ForestGreen}{0.001927} & 0.369116 & \textcolor{ForestGreen}{0.044758} & 0.048970 & 0.020815 & 12.259072 & 1.741011 & \textcolor{blue}{0.0351} & \textcolor{ForestGreen}{0.1031} & 0.0260 & 0.0463 \\
     AIVS & 0.000502 & 0.004217 & 0.518831 & 0.080380 & \textcolor{blue}{0.002092} & \textcolor{red}{0.016377} & 9.786209 & 1.754694 & 0.0321 & 0.0760 & 0.0142 & \textcolor{blue}{0.0530}  \\
          Potamias et al. & 0.000240 & 0.002117 & \textcolor{blue}{0.327904} & \textcolor{red}{0.012680} & 0.002247 & 0.028753 & \textcolor{blue}{6.091248} & \textcolor{red}{0.669750} & 0.0538 & 	
\textcolor{red}{0.1174} & \textcolor{ForestGreen}{0.0376} & \textcolor{ForestGreen}{0.0606}  \\
    \bottomrule
    \\
    \end{tabular}
    }
    \label{tab:results_extra}
\end{table*}


\begin{table*}[!]
\vspace{-1cm}
    \caption{Ablation and noise studies on the Stanford bunny and Armadillo respectively. Here,  $k_{init}$, $i$, $k_{opt}$, $k$, and $n$ are the number of initial points chosen using FPS, the number of times the hyperparameters of the GP are optimized, the number of points used for hyperparameter optimization, the number of neighbourhood points used for surface variation estimation, and the noise factors respectively (Section \ref{ablation_sec}).}
    \centering
    \resizebox{\linewidth}{!}{
    \begin{tabular}{cccccccccc}
    \toprule 
    \multicolumn{8}{c}{Ablation studies} & \multicolumn{2}{c}{Noise studies} \\
    \cmidrule(lr){0-7} \cmidrule(lr){9-10}
        $k_{init}$ & CD & $i$ & CD & $k_{opt}$ & CD & $k$ & CD & $n$ & CD\\
        \midrule
     500 & 7.4345e-06 & 50 & 3.1976e-06 & 50 & 3.2453e-06 & 5 & 2.7721e-06 & 0.0001 & 1.9456\\
     1000 & 5.0463e-06 & 100 &  3.2049e-06 & 100 & 3.2237e-06 & 10 & 3.0523e-06 & 0.0005 & 1.9532\\
     1500 &  3.8995e-06 & 150 & 3.2471e-06 & 150 & 3.2156e-06 & 15 & 3.1314e-06 & 0.001 & 1.9148\\
     2000 & 3.2469e-06 & 200 & 3.2153e-06 & 200 &  3.2529e-06 & 20 & 3.2396e-06 & 0.005 & 2.1729\\
     3000 &  2.4893e-06 & 300 & 3.2253e-06 & 300 & 3.2139e-06 & 30 & 3.3033e-06 & 0.01 & 3.1361\\
      5000 & 1.8498e-06 & 500 & 3.1996e-06 & 500 & 3.2236e-06 & 50 & 3.4406e-06 & 0.02 & 4.8644\\
    \bottomrule
    \\
    \end{tabular}
    }
    \label{tab:ablation}
\end{table*}

\begin{figure*}
 \hspace{-1cm}
        \includegraphics[trim=0cm 0cm 0cm 0cm, clip, scale=0.9]{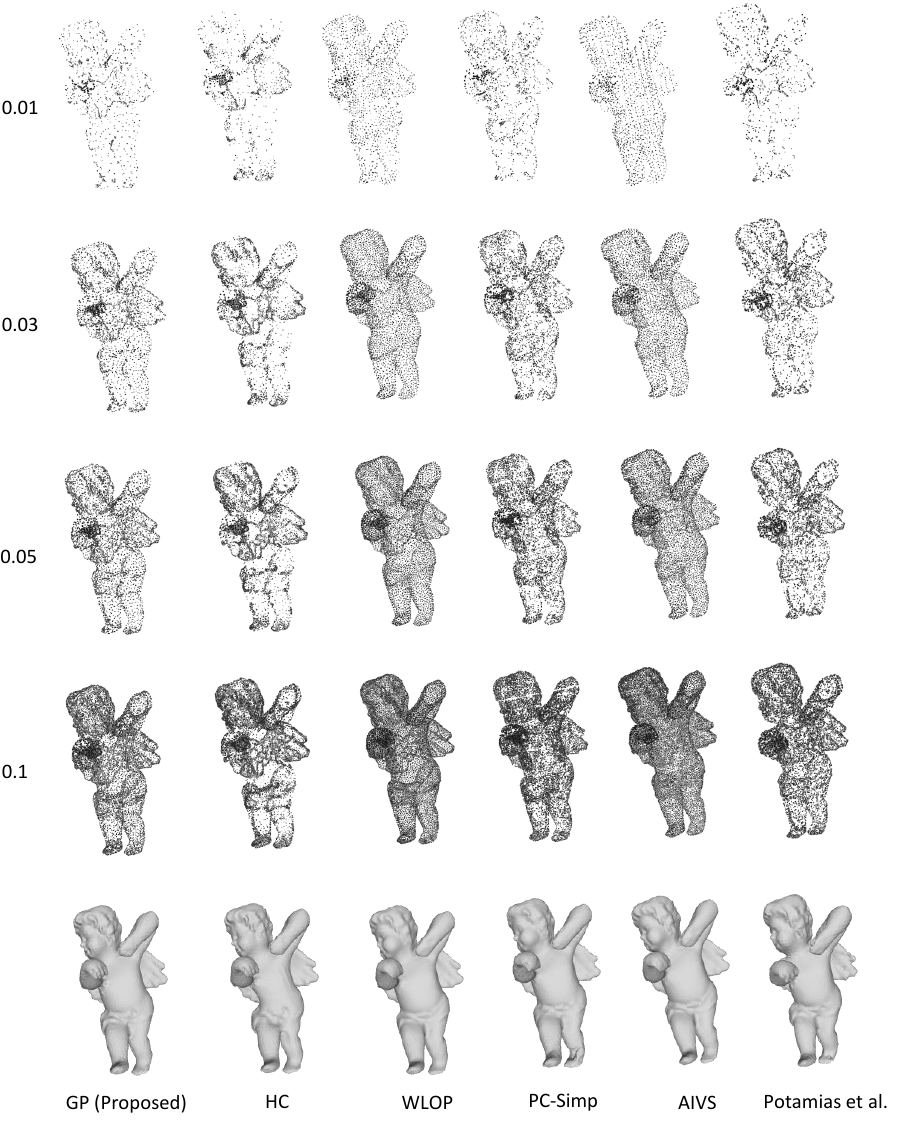}
         \caption{Simplified representations of the angel point cloud for a variety of simplification ratios (leftmost column) and the meshes reconstructed from clouds obtained using simplification ratio $\alpha=0.05$ for all evaluated simplification techniques (last row).}
         \label{fig:angel}
\end{figure*}

\end{document}